\documentclass[runningheads]{llncs}

\usepackage[T1]{fontenc}
\usepackage[utf8]{inputenc}
\usepackage{graphicx}
\usepackage{booktabs}
\usepackage{multirow}
\usepackage{amsmath}
\usepackage{amssymb}
\usepackage{url}
\usepackage{float}
\usepackage{placeins}
\usepackage{microtype}
\usepackage{enumitem}
\setlist{nosep}

\begin{document}

\title{Interpretable Policy Distillation for Power Grid Topology Control}
\titlerunning{Interpretable Policy Distillation for Power Grid Control}
\author{Aleksandra Dmitruka\inst{1} \and K\={a}rlis Freivalds\inst{1}}
\authorrunning{A. Dmitruka and K. Freivalds}
\institute{$^1$University of Latvia, Faculty of Exact Sciences and Technology, Riga, Latvia\\
\email{ad18165@students.lu.lv}
\email{karlis.freivalds@lu.lv}}

\maketitle

\begin{abstract}
Deep reinforcement learning (RL) offers a promising route to real-time power grid operation, yet large neural policies are costly to evaluate, hard to deploy on constrained hardware, and opaque to operators. We ask whether a Proximal Policy Optimization (PPO) agent for grid topology control can be compressed into compact tree-based surrogates without losing operational performance. A PPO teacher is trained on Grid2Op's standard 14-bus environment with a stability-oriented reward, using stress-focused data collection on critical, high-loading states. The policy is then distilled into a decision tree and a random forest. Across held-out validation episodes, both surrogates exceed the teacher in mean reward and survival length at a fraction of the inference cost. The decision tree shows high exact-action agreement with the PPO argmax and near-complete agreement within its top-ranked actions, while remaining small enough to be inspected directly. Feature-importance analysis reveals a representational shift: the PPO policy relies mainly on line-loading signals, while the distilled tree is driven primarily by bus-topology variables. These results suggest that stress-focused distillation can convert a black-box neural controller into a lightweight, auditable rule-like surrogate suited for real-time deployment, while also surfacing risks tied to deterministic actions and topology-specific generalization.

\end{abstract}

\section{Introduction}

Power transmission systems are operated ever closer to their physical limits, as renewable generation, rising demand, and decarbonization push operators toward faster decisions under uncertainty. Topology control---switching grid elements between busbars or reconnecting lines to alter feasible power-flow paths---is attractive in this setting because it can relieve line overloads without curtailing generation or disconnecting consumers. A useful action, however, must be chosen quickly, must remain robust under stress, and must be one that engineers can inspect and challenge.

Reinforcement learning has received attention as a tool for power network operation \cite{kelly2020reinforcement,yoon2021winning,lehna2024hugo}, and platforms such as Grid2Op make this research reproducible \cite{grid2opdoc}. RL policies can learn non-trivial reactions to overloads and contingencies, yet the properties that make them attractive also create a deployment barrier. A black-box controller is hard to trust when an action may trigger cascading failures, violate operational constraints, or be hard to justify afterward, and evaluating a large neural policy inside a real-time control loop adds further cost. Both opacity and inference cost fall in the scope of efficient deep learning, and in high-stakes domains inherently interpretable models are increasingly argued to be the preferable choice when performance is comparable \cite{rudin2019stop,molnar2020interpretable}.

This paper investigates a middle ground: train a neural RL controller, then distil it into compact, interpretable tree-based models. The teacher is an MLP PPO agent trained in Grid2Op; the students are a decision tree and a random forest trained over its discrete actions. The neural network is used during training to explore the action space, while the deployed controller is smaller, faster to inspect, and easier to constrain with safety filters. The main question is whether the students remain operationally useful in closed-loop simulation: a student may agree with the teacher in aggregate yet fail on rare safety-critical cases.

The work treats interpretable policy distillation as an efficient deep-learning method for safety-critical control. Its contribution is threefold. First, it presents an end-to-end pipeline for PPO training, stress-state collection, and tree-based distillation in the \texttt{l2rpn\_case14\_sandbox} environment. Second, it compares teacher and students using both imitation and closed-loop metrics: cumulative reward, survival length, completed-week milestones, and execution time. Third, it analyzes interpretability through direct tree inspection and permutation feature importance, showing that the distilled policy can rely on different input features than the teacher while preserving or improving performance.

\section{Background and Related Work}

\subsection{Power Grid Topology Control}

A transmission grid consists of substations, busbars, lines, generators, and loads. Its operating state is constrained by thermal limits, voltage limits, and N-1 security criteria. The relative line loading ratio, denoted by $\rho$, is central to the Grid2Op setting: values close to one indicate that a line is near its thermal limit, and persistent overloads may cause disconnection. To relieve such overloads, operators can use topology control, which alters substation connectivity and redistributes flows across the network. This lever is operationally useful but combinatorial: a small benchmark can already produce a large discrete action space.

The Learning to Run a Power Network (L2RPN) benchmark family formalized this problem as sequential decision-making with realistic power-flow simulation \cite{kelly2020reinforcement,yoon2021winning}. Recent work has explored action-space reduction, graph-based RL architectures, heuristic target topologies, and benchmarking methodology for RL-based grid operation \cite{zhou2021action,batanero2025graph,marchesini2025rl2grid,lehna2024hugo}. Together, these studies show that RL controllers can be effective on such benchmarks. A separate challenge for high-stakes deployment is that their recommendations must also be auditable by human operators \cite{rudin2019stop}.

\subsection{Reinforcement Learning and Reward Design}
The grid-control problem can be represented as a Markov decision process with state $s_t$, action $a_t$, transition dynamics, reward $r_t$, and discount factor $\gamma$ \cite{sutton2018reinforcement}. The agent learns a policy $\pi(s)$ that maps states to actions, aiming to maximize the expected cumulative discounted sum of rewards over a horizon $H$. Following \cite{hassouna2026graph}, the optimal policy $\pi^*$ is then
\begin{equation}
\pi^{*} = \arg\max_{\pi}\;\mathbb{E}\!\left[\sum_{t=0}^{H}\gamma^{t}\,r(s_t, a_t)\right].
\end{equation}
Reward design is decisive: shaping can guide the agent toward safer behaviour but may also distort the intended objective \cite{ng1999policy}. In power grids, a reward should encourage survival and spare line capacity while penalizing overloads, illegal actions, and blackouts.

\subsection{Knowledge Distillation and Interpretability}
Knowledge distillation trains a simpler student to reproduce the behaviour of a stronger teacher \cite{hinton2015distilling}, using hard labels, soft probabilities, or intermediate representations. In RL, policy extraction and distillation convert a learned policy into an interpretable or verifiable model that can serve as a controller, safety monitor, or explanation layer \cite{bastani2018verifiable,coppens2019distilling,vos2024optimizing}.

Among interpretable student models, decision trees are attractive because their decisions are sequences of explicit threshold tests. Random forests, by contrast, improve robustness through variance reduction, but the ensemble structure weakens direct interpretability \cite{breiman2001random}. To analyze both, this work combines ante-hoc interpretability (direct tree inspection) with post-hoc tools such as permutation feature importance, which can compare the signals used by different policies but should be interpreted cautiously when features are correlated \cite{molnar2020interpretable}.

\section{Experimental Environment}

The experiments use Grid2Op with the \texttt{l2rpn\_case14\_sandbox} environment. The environment is based on a modified IEEE 14-bus grid and contains 14 substations, 20 transmission lines, 6 generators, and 11 loads. It contains chronological episodes (``chronics'') describing changing consumption and generation. The full dataset contains 1004 episodes. An 80/20 split is used: 804 episodes for training and data collection, and 200 episodes reserved as a validation pool. The closed-loop evaluation reported later uses 100 held-out validation episodes, each capped at 6048 steps. With five-minute steps, this corresponds to approximately three weeks per episode.

The observation vector includes nine groups of variables: line loading $\rho$, generator active power, load active power, topology vector, line status, time steps before overflow disconnection, cooldown time for lines, cooldown time for substations, and actual dispatch. The action space contains 219 discrete topology-related actions. A DoNothing agent is used as the baseline.

\section{Methodology}

\subsection{Teacher Training}

The PPO teacher is trained through the Gymnasium-compatible interface \cite{gymnasium2024}. Two custom wrappers shape the learning process. The first, \emph{StabilityActionReward}, rewards safe operation, spare line capacity, and longer survival, while penalizing overloads, illegal actions, and system failure. The second, \emph{SafeStateSkipper}, skips stable periods so training focuses on stressed states where decisions matter.

The PPO configuration uses a discount factor $\gamma=0.99$, learning rate $3\times10^{-4}$, entropy coefficient 0.01. The policy/value architecture is an MLP with two hidden layers of 256 neurons. Training is run for 200,000 environment steps.

\begin{figure}[!htbp]
\centering
\includegraphics[width=.7\textwidth]{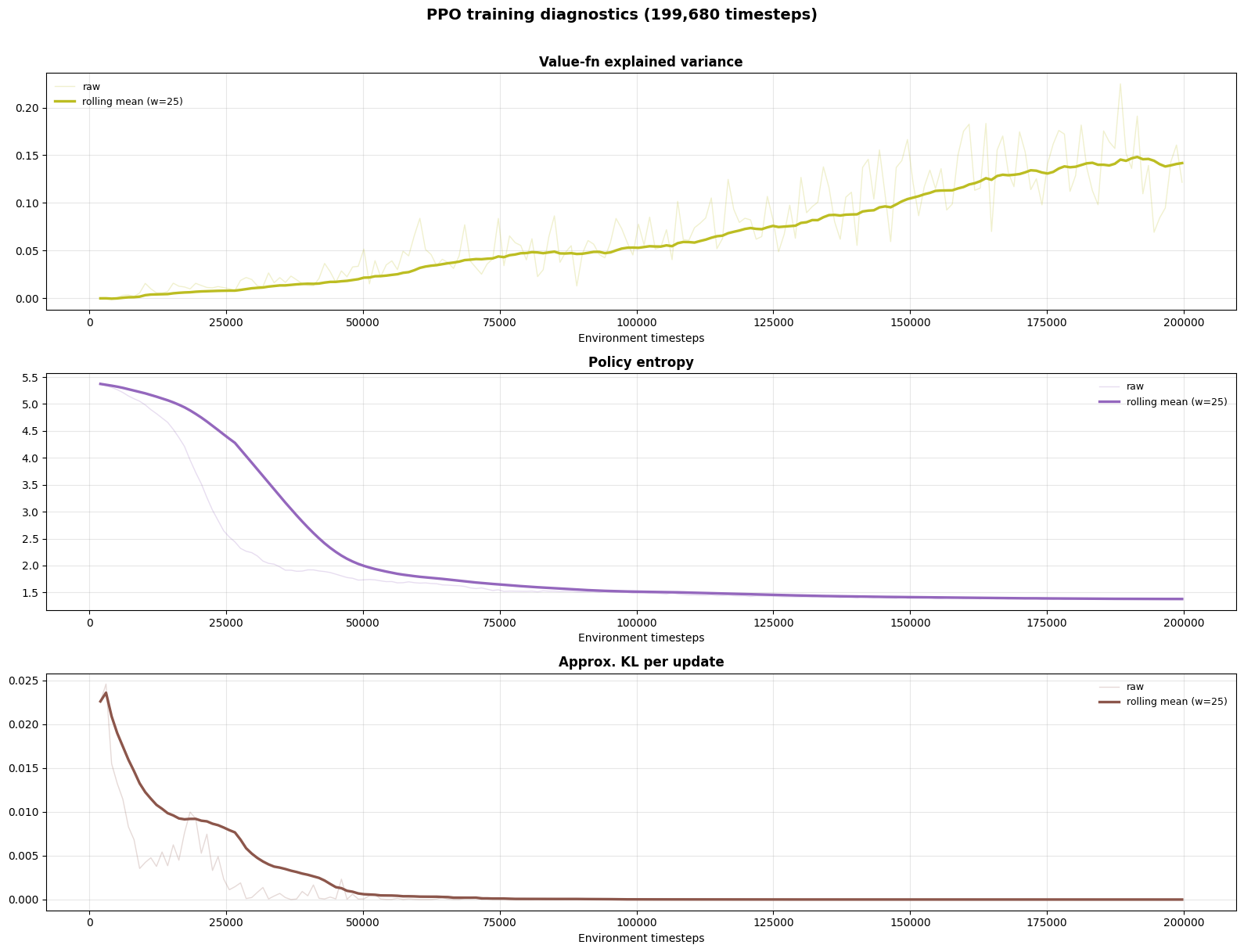}
\caption{PPO training diagnostics over 200,000 environment steps.}
\label{fig:train}
\end{figure}

Figure~\ref{fig:train} summarizes the training diagnostics. The explained variance of the value function grows from near zero to about 0.15 over training. Policy entropy falls from roughly 5.35 to 1.40, indicating partial but not complete determinization. Approximate KL divergence drops sharply after about 50,000 steps, suggesting early stabilization or convergence to a local optimum. The teacher is therefore useful but not necessarily fully optimized.

\vspace{-0.5em}
\vspace{-1em}

\subsection{Stress-Focused Policy Distillation}
After training, the PPO agent is used as a teacher to generate a supervised distillation dataset. Rather than recording every encountered state, the procedure stores only critical or pre-critical states for which at least one line has $\max(\rho)>0.85$. This threshold lies deliberately below the active intervention region near $\rho=0.9$, so the student sees states where intervention is becoming relevant rather than only states that are already unsafe.

\begin{figure}
\centering
\includegraphics[width=.92\textwidth]{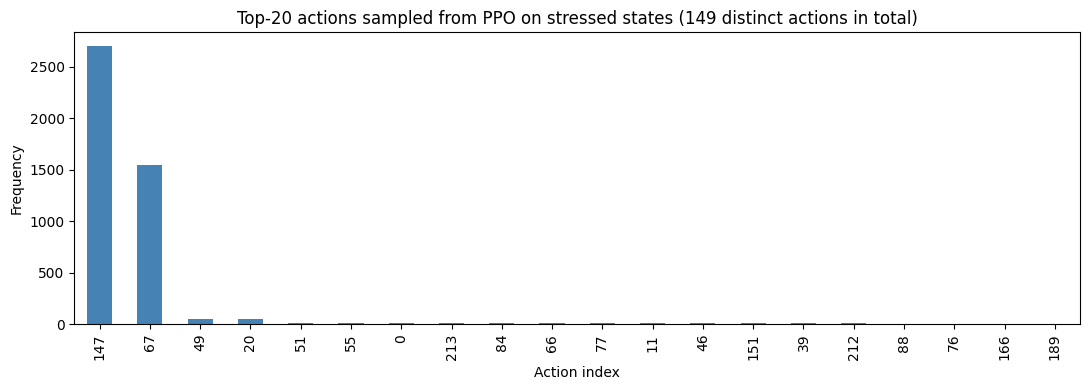}
\caption{Twenty most frequent PPO actions in critical grid states.}
\label{fig:actions}
\end{figure}

{\setlength{\parskip}{0pt}
For each selected state, three state--action samples are stored, with actions sampled from the PPO policy distribution rather than always taking the argmax. This captures policy uncertainty but introduces label noise, as the same observation may appear with different target actions. The resulting multiclass classification task uses the same normalized observation vector as the PPO teacher, with the discrete action index as target.

Two student models are trained with scikit-learn: a decision tree (capped at 100 leaves) as the directly interpretable model, and a random forest (100 trees) as a stronger tree-based comparison that can model richer boundaries at the cost of inspectability. Both are evaluated through imitation metrics on a test set (exact, top-3, and top-5 argmax agreement) and closed-loop simulation as wrapped Grid2Op agents.

The teacher's action use is highly concentrated (Fig.~\ref{fig:actions}). Out of 219 possible actions, the teacher uses 149 in the critical-state dataset, and two dominate: action 147 (selected $\sim$2700 times) and action 67 ($\sim$1540 times). Both are topology-only bus-assignment actions: action 147 reconfigures substation 8, and action 67 reconfigures substation 3. This concentration helps explain why a compact student can succeed -- it mainly has to decide \emph{when} to apply a small set of recurrent substation reconfigurations, not reproduce a uniformly complex policy over the full action space.
\par}

\vspace*{-4.5em}

\section{Results}
\vspace{-3em}
\subsection{Compression Ratio and Imitation Fidelity}
{
\setlength{\intextsep}{6pt plus 2pt minus 2pt}
\setlength{\textfloatsep}{6pt plus 2pt minus 2pt}
\setlength{\floatsep}{6pt plus 2pt minus 2pt}
\setlength{\abovecaptionskip}{4pt}
\setlength{\belowcaptionskip}{4pt}
\vspace{-3.4em}

Table~\ref{tab:structure} reports structural complexity. The decision tree has depth 18, 100 leaves, and 199 nodes. The random forest consists of 100 trees with average depth 35.5, average 953.6 leaves per tree, and 190,628 total nodes. This is a large interpretability gap: a single tree can be inspected and converted to rule-like logic, whereas a forest of this size is mainly interpretable through aggregate tools.

\vspace{6em}
\begin{table}[H]
\centering
\caption{Structural characteristics of distilled models.}
\label{tab:structure}
\setlength{\tabcolsep}{10pt}
\renewcommand{\arraystretch}{1.25}
\begin{tabular}{lrrr}
\toprule
Model & Depth & Leaves & Nodes \\
\midrule
Decision tree & 18 & 100 & 199 \\
Random forest, mean per tree & 35.5 & 953.6 & 1906 \\
Random forest, total & \multicolumn{1}{c}{--} & \multicolumn{1}{c}{--} & 190628 \\
\bottomrule
\end{tabular}
\end{table}
\vspace{1em}

Table~\ref{tab:agreement} gives imitation accuracy. The decision tree agrees with the PPO argmax in 78.3\% of test states and reaches 99.9\% top-3/top-5 agreement. The random forest performs worse on all three measures: 63.3\% argmax, 93.8\% top-3, and 94.7\% top-5. This is counterintuitive because the forest is more expressive. The likely explanation is that the distillation labels are stochastic: with three samples per state from the PPO distribution, the forest may smooth over inconsistent labels, whereas the constrained tree locks onto the dominant modes more sharply.

\vspace{1em}
\begin{table}[H]
\centering
\caption{Agreement of student actions with the PPO teacher on the test set.}
\label{tab:agreement}
\setlength{\tabcolsep}{12pt}
\renewcommand{\arraystretch}{1.25}
\begin{tabular}{lrrr}
\toprule
Model & Argmax & Top-3 & Top-5 \\
\midrule
Decision tree & 78.3\% & 99.9\% & 99.9\% \\
Random forest & 63.3\% & 93.8\% & 94.7\% \\
\bottomrule
\end{tabular}
\end{table}
\vspace{1em}

Figure~\ref{fig:tree} shows the upper levels of the learned decision tree. The visible splits are primarily topology conditions, such as whether a given element is connected to a specific bus. This is operationally meaningful: the model's reasoning can be described as a sequence of busbar configuration checks, not as an opaque activation pattern.

\vspace{0.3em}
\begin{figure}[H]
\centering
\includegraphics[width=.98\textwidth]{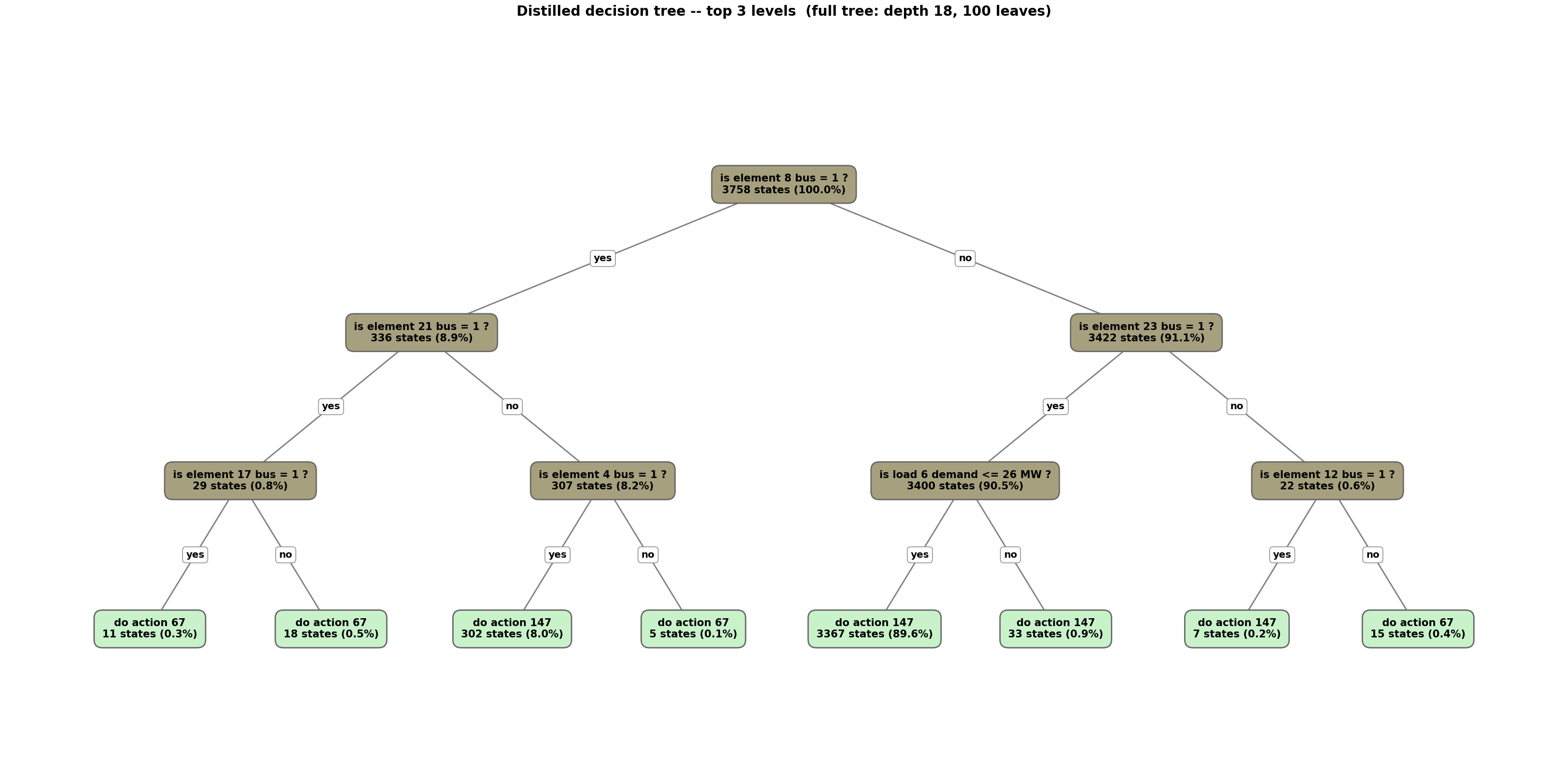}
\caption{Top levels of the distilled decision tree.}
\label{fig:tree}
\end{figure}
\vspace{0.5em}

}
\FloatBarrier

The upper tree confirms that the first rule is not an abstract numerical threshold but a topology question about the bus assignment of element 8. The largest visible leaf covers 89.6\% of the samples and predicts action 147, while most smaller leaves predict action 67. Non-topological thresholds (such as load-demand splits) appear only at deeper levels. The distilled policy is therefore more than a black-box approximation: its dominant behaviour can be expressed as a small set of substation-configuration checks leading to two concrete bus-switching actions.

\subsection{Closed-Loop Operational Performance}
Operational performance is evaluated over 100 held-out validation episodes. Table~\ref{tab:compute} reports total executed steps and wall-clock time. The DoNothing baseline terminates early in many episodes and therefore executes only 158,437 total steps. PPO executes 388,372 steps. The two distilled agents execute approximately 472,800 steps each, about 22\% more than PPO. Their longer wall-clock time is therefore not simply a slower decision function; they survive longer and consequently make more decisions.

\vspace{-0.5em}
\begin{table}[H]
\centering
\caption{Closed-loop computational cost over 100 held-out validation episodes.}
\label{tab:compute}
\setlength{\tabcolsep}{10pt}
\renewcommand{\arraystretch}{1.25}
\begin{tabular}{lrrr}
\toprule
Agent & Total steps & Total time (s) & Mean time per episode (s) \\
\midrule
DoNothing & 158437 & 606.9 & 6.1 \\
PPO & 388372 & 1222.4 & 12.2 \\
Decision tree & 472802 & 1416.9 & 14.2 \\
Random forest & 472804 & 1489.5 & 14.9 \\
\bottomrule
\end{tabular}
\end{table}
\vspace{-0.5em}

Table~\ref{tab:survivalweeks} shows survival milestones. Only 3 DoNothing episodes reach the full three-week cap. PPO reaches the cap in 37 episodes. Both distilled models reach it in 59 episodes, improving over PPO by 22 episodes and over DoNothing by 56 episodes. The one-week and two-week counts show the same ordering.

\vspace{-0.5em}
\begin{table}[H]
\centering
\caption{Number of episodes reaching one-, two-, and three-week milestones.}
\label{tab:survivalweeks}
\setlength{\tabcolsep}{12pt}
\renewcommand{\arraystretch}{1.25}
\begin{tabular}{lrrr}
\toprule
Agent & 1 week & 2 weeks & 3 weeks \\
\midrule
DoNothing & 26 & 8 & 3 \\
PPO & 72 & 53 & 37 \\
Decision tree & 85 & 72 & 59 \\
Random forest & 85 & 72 & 59 \\
\bottomrule
\end{tabular}
\end{table}
\vspace{-0.5em}

Reward and survival-step metrics are reported in Tables~\ref{tab:reward} and~\ref{tab:steps}. PPO improves strongly over DoNothing, but both students improve further. The decision tree reaches a mean reward of 5583.961 compared with 4529.932 for PPO and 1821.459 for DoNothing, a 23.3\% gain over the teacher and a 206.6\% gain over the baseline. The random forest is almost identical to the decision tree. Median reward is also higher for the students, and the decision tree's reward standard deviation is 9.5\% lower than PPO's. For survival length, the students reach a mean of approximately 4728 steps and a median of 6048, meaning that at least half of their episodes reach the maximum length. Their mean survival is 21.7\% higher than PPO's, and their three-week completion count is 59 rather than 37 episodes, a 59.5\% relative improvement.
\vspace{-0.2em}

\vspace{-0.2em}
\begin{table}[!htbp]
\centering
\caption{Reward metrics over 100 held-out validation episodes.}
\label{tab:reward}
\setlength{\tabcolsep}{8pt}
\renewcommand{\arraystretch}{1.25}
\begin{tabular}{lrrrr}
\toprule
Agent & Mean reward & Median reward & Std. reward & Min. reward \\
\midrule
DoNothing & 1821.5 & 1238.2 & 1627.1 & -9.0 \\
PPO & 4529.9 & 5321.5 & 2489.7 & 589.9 \\
Decision tree & 5584.0 & 7109.9 & 2253.3 & 595.7 \\
Random forest & 5580.6 & 7109.7 & 2251.9 & 595.7 \\
\bottomrule
\end{tabular}
\end{table}
\vspace{-3em}

\begin{table}[!htbp]
\centering
\caption{Survival-step metrics over 100 held-out validation episodes, with a maximum of 6048 steps per episode.}
\label{tab:steps}
\setlength{\tabcolsep}{10pt}
\renewcommand{\arraystretch}{1.25}
\begin{tabular}{lrrr}
\toprule
Agent & Mean steps & Median steps & Max. steps \\
\midrule
DoNothing & 1584.4 & 1087 & 6048 \\
PPO & 3883.7 & 4549 & 6048 \\
Decision tree & 4728.0 & 6048 & 6048 \\
Random forest & 4728.0 & 6048 & 6048 \\
\bottomrule
\end{tabular}
\end{table}
\vspace{-0.5em}

Taken together, the tables show that distillation does not merely approximate the teacher; in these evaluation runs, it improves both reward and survival while reducing reward dispersion. The lower standard deviation in particular suggests that the students behave more consistently across episodes, not just better on average.

\begin{figure}[t]
\centering
\includegraphics[width=.8\textwidth]{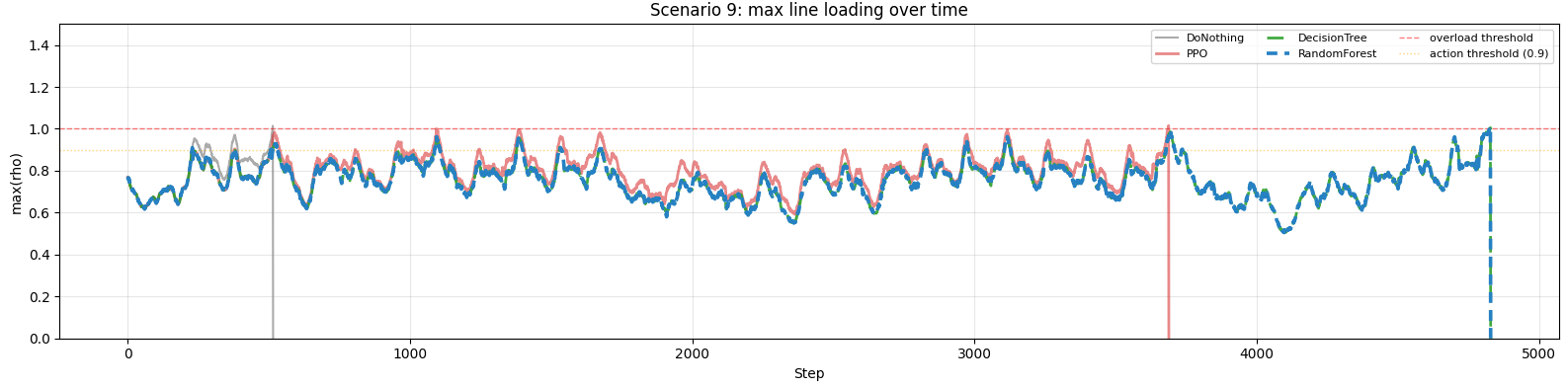}\vspace{1mm}
\includegraphics[width=.8\textwidth]{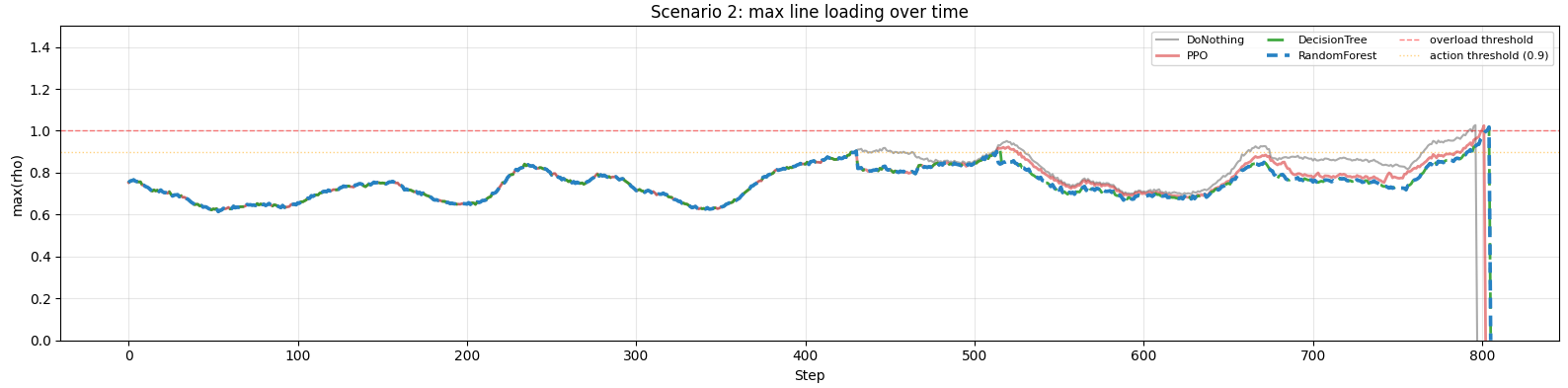}\vspace{1mm}
\includegraphics[width=.8\textwidth]{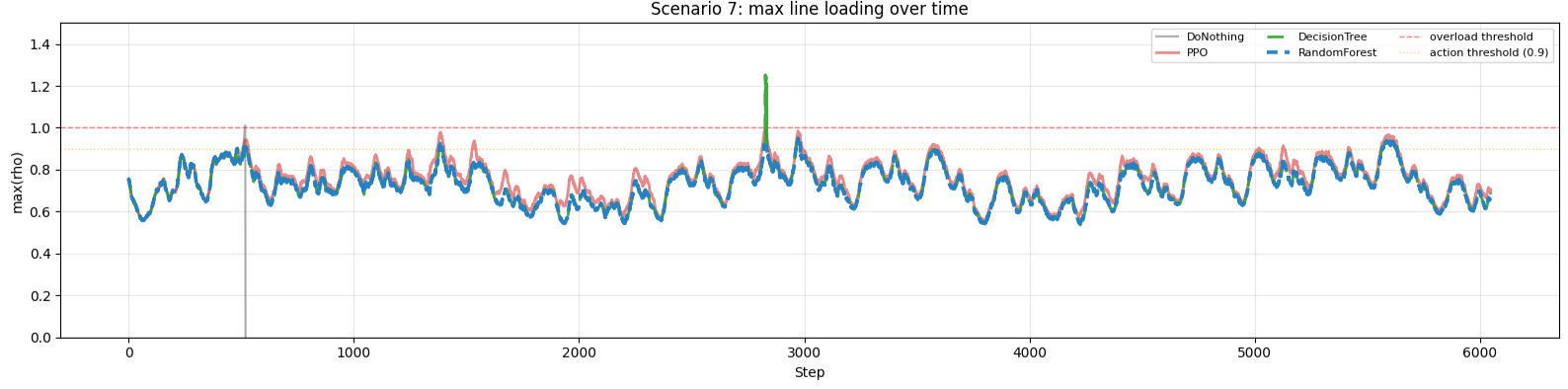}
\caption{Maximum line loading $\max(\rho)$ in three representative validation scenarios.}
\label{fig:rho}
\end{figure}

Scenario-level line-loading plots provide a more cautious view (Fig.~\ref{fig:rho}). In scenario 9, DoNothing collapses around step 500 as a single overloaded line cascades, PPO survives to roughly 3700 before a similar failure, and both distilled models survive to about 4800. In scenario 2, all four agents fail around the same time, suggesting that some cases are not solved by the current action/reward design and may require either richer action sets or longer-horizon planning. In scenario 7, the distilled models finish the full episode, but the decision tree briefly drives one line above $\rho=1.2$. Grid2Op allows temporary overloads, so this is not automatically a failure, but it indicates a reduced safety margin and shows that high cumulative reward can coexist with locally unsafe transients.

\subsection{Feature Importance and Policy Explanation}
The top-level decision-tree visualization already suggests that topology variables dominate the distilled decision logic. The same pattern emerges from internal tree Gini importance (reported in the appendix): 14 of the top 20 features are topology variables, while only two are direct line-loading variables. Because internal importance is model-specific, permutation feature importance is then used to compare PPO, the decision tree, and the random forest on a common basis.

\begin{figure}[t]
\centering
\includegraphics[width=.42\textwidth]{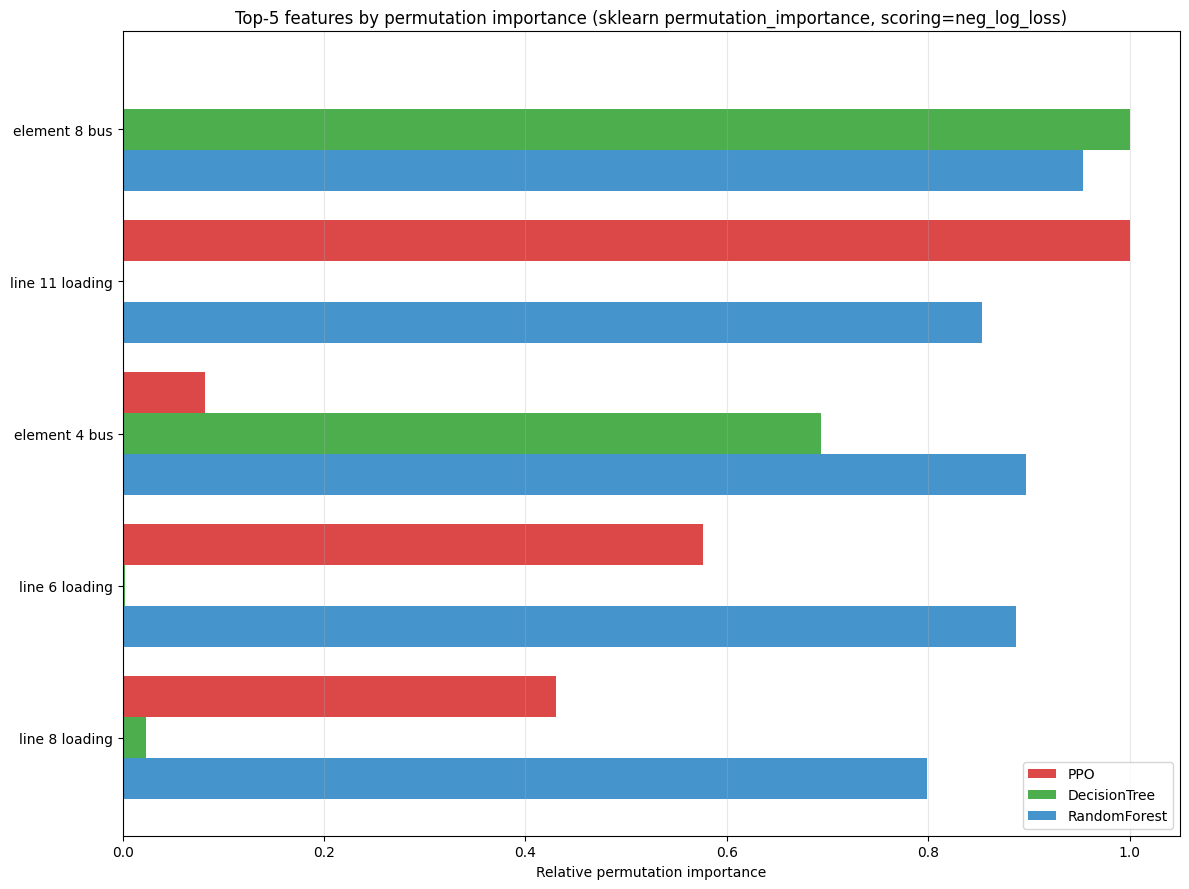}\hfill
\includegraphics[width=.58\textwidth]{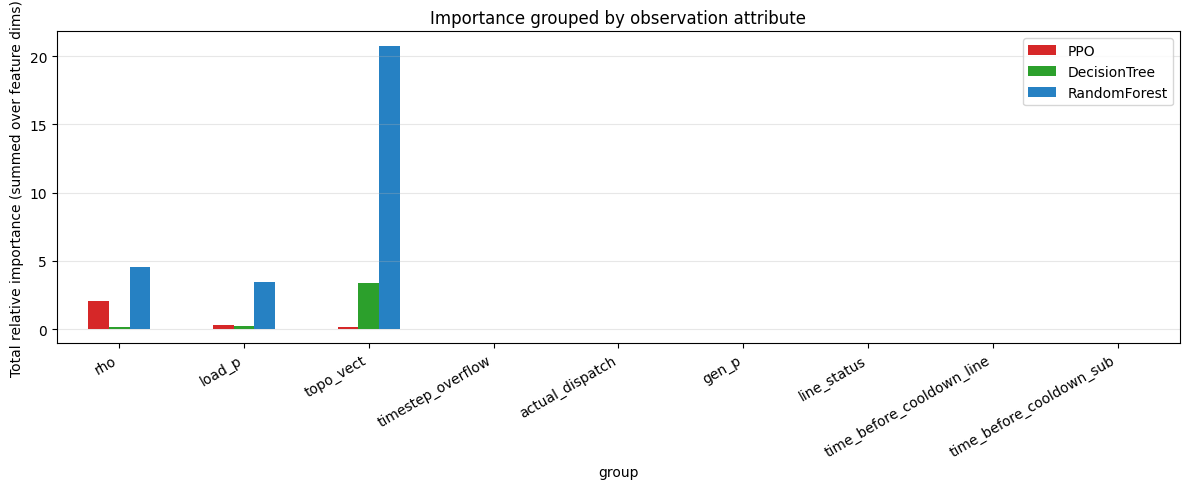}
\caption{Permutation feature importance by individual features (left) and by Grid2Op observation groups (right).}
\label{fig:importance}
\end{figure}

Figure~\ref{fig:importance} shows the five most important individual features and grouped importances by observation type. The PPO teacher assigns the highest importance to line 11 loading, whereas both students place their top feature on the bus assignment of element 8. In the grouped view, bus topology dominates the students while line loading dominates PPO. The students therefore approximate the teacher's action choices through a different representation of the grid state.

This difference is not necessarily a flaw. Operators reason about switching in topological terms -- which element is on which bus, and which substation configuration should change -- so a tree expressed in those terms may be easier to audit than a controller citing only continuous loading values. At the same time, topology-dominant rules may generalize poorly if the same topology appears with different electrical conditions. This is the central trade-off found in the experiment.

\FloatBarrier

\section{Discussion}

\subsection{Why Can the Students Outperform the Teacher?}
Several mechanisms make this plausible. First, the teacher is only partially converged: final entropy remains around 1.40 and value-function explained variance well below one, so its stochastic policy can pick suboptimal actions in states where several actions have similar probability. The students, by contrast, are deterministic. Second, the distillation dataset is filtered to $\max(\rho)>0.85$, so stable states do not dominate the supervised objective and the students spend their capacity on decisions that matter operationally. Third, tree-based models naturally exploit discrete topology variables, which in this environment appear to contain a compact signal about useful switching decisions -- a representation that may be simpler and more effective than the teacher's reliance on continuous line-load signals.

This should not be overgeneralized: longer training or graph-based architectures may change the outcome. The conclusion is narrower but still useful: a neural policy can produce enough behavioural data to train a compact controller that is both interpretable and operationally competitive.

\subsection{Interpretability versus Safety}
The decision tree is the most attractive practical artefact in this study: compact, high top-3 agreement with the teacher, closed-loop performance matching the random forest, and directly inspectable -- its first decisions are bus-topology tests that translate into operational rules. This makes it a candidate for decision support, offline validation, or a human-in-the-loop recommendation system.

The same determinism that makes the tree easy to audit can create risk. The scenario 7 overload spike above $\rho=1.2$ shows that small state variations may not be represented finely enough, and the tree always maps the same input to the same action. A real deployment would need guardrails: hard action filters, contingency checks, uncertainty detection, and fallback to conservative operation when out of distribution. The tree should therefore be treated as an interpretable controller candidate, not a safety-certified system. The random forest does not justify its complexity here: it matches the decision tree in closed-loop metrics but has lower imitation agreement and vastly larger structure. Its main value is diagnostic, confirming that the performance gain is not exclusive to one particular tree. For operator-facing use, the single tree is preferable.

\subsection{Limitations}

The study is limited to \texttt{l2rpn\_case14\_sandbox}, which is useful for interpretability but does not prove scalability to large transmission networks. The evaluation covers 100 held-out episodes capped at 6048 steps; larger scenario sets and adversarial contingencies would be needed for stronger claims. The distillation dataset is also biased toward stressed states above a fixed $\rho$ threshold; this bias improves performance here, but a different threshold could change the balance between early intervention and unnecessary switching.

Imitation uses sampled teacher actions rather than soft-label distillation, which introduces contradictory labels; future work should compare hard, soft, and risk-weighted targets. Formal verification is also absent: safety constraints must be enforced independently of the learned model.

\section{Conclusion}

This paper presented a policy-distillation workflow for efficient, interpretable RL in power-grid topology control. A PPO teacher was trained in Grid2Op and distilled into a decision tree and a random forest on critical states with $\max(\rho)>0.85$. Across 100 held-out validation episodes, both students outperformed PPO and the DoNothing baseline. The decision tree reached 78.3\% exact and 99.9\% top-3 agreement with the PPO argmax, a mean reward of 5584 (vs.\ 4530 for PPO), and a median survival of 6048 steps (vs.\ 4549).

The main implication is that policy distillation can be used as a deployment-oriented compression step for neural control policies. Rather than treating a deep RL model as the final artefact, it can be treated as a training-time teacher whose behavior is transferred to a simpler controller. This is especially relevant for real-time and resource-aware AI systems, where inference transparency, validation cost, and operational robustness matter as much as raw benchmark performance.

\begin{credits}
\subsubsection{Acknowledgements}
The authors would like to thank Rebeka Birziņa, a programming engineer at the Institute of Electronics and Computer Science, for her insight into reinforcement learning solutions in the field of power grids, as well as for her valuable support in identifying the most relevant research directions.

\subsubsection{Disclosure of Interests} The authors have no competing interests to declare that are relevant to the content of this article.
\end{credits}

\bibliographystyle{splncs04}
\bibliography{refs}

\end{document}